\newif\ifreview
\reviewfalse  

\documentclass[10pt,conference]{IEEEtran}
\usepackage{balance}
\usepackage{tabularx}
\usepackage{booktabs}
\usepackage{adjustbox}
\usepackage{xurl}
\usepackage{hyperref}

\usepackage{enumitem}
\newlist{rqlist}{enumerate}{1}
\setlist[rqlist]{label*=\textbf{RQ\arabic*}, ref=RQ\arabic*}

\newcommand{\python}{{\path{Python}}}
\newcommand{\qwlil}{{\path{Qwen3-8B}}}
\newcommand{\qwmid}{{\path{Qwen3-14B}}}
\newcommand{\qwbig}{{\path{Qwen3-32B}}}
\newcommand{\llama}{{\path{Llama-R1-8B}}}
\newcommand{\olmo}{{\path{OLMo3-7B}}}
\newcommand{\openr}{{\path{OpenReason-7B}}}
\newcommand{\opencr}{{\path{CodeReason-7B}}}

\newcommand{\evalplus}{{\path{EvalPlus}}}
\newcommand{\livecodebench}{{\path{LiveCodeBench}}}
\newcommand{\codecontests}{{\path{CodeContests}}}
\newcommand{\bigcodebench}{{\path{BigCodeBench}}}
\newcommand{\crux}{{\path{CRUXEval}}}
\newcommand{\cruxin}{{\path{CRUX-Input}}}
\newcommand{\cruxout}{{\path{CRUX-Output}}}

\usepackage[most]{tcolorbox}
\usepackage[dvipsnames,table]{xcolor}
\definecolor{boxborder}{HTML}{009400}
\definecolor{boxinner}{HTML}{F2F2F2}

\definecolor{summarybg}{RGB}{242,247,252}
\definecolor{summaryborder}{RGB}{0,0,102}

\newcommand{\weburl}[1]{
\href{#1}{\texttt{\textcolor{summaryborder}{#1}}}
}

\usepackage{fancyhdr}

\ifreview
    \newcommand{\repourl}{{\weburl{https://anonymous.4open.science/r/546B}}}

\else
    \newcommand{\repourl}{{\weburl{https://github.com/itsluketwist/think-overflow}}}
\fi

\newtcolorbox{summarybox}[1]{
  colback=summarybg,
  colframe=summaryborder,
  colbacktitle=summaryborder,
  coltitle=white,
  title=\textsc{#1},
  fonttitle=\bfseries,
  arc=4pt,
  boxrule=1pt,
  left=6pt,
  right=6pt,
  top=6pt,
  bottom=6pt
}

\begin{document}
%
\title{Reasoning-Trace Overflow in Code Generation:\\An Empirical Study of Explicit Reasoning Models}

\ifreview
\author{
\IEEEauthorblockN{Anonymous Authors}
\IEEEauthorblockA{
\\
}
}

\else

\author{
\IEEEauthorblockN{Lukas Twist}
\IEEEauthorblockA{
King's College London\\
London, UK}
\and
\IEEEauthorblockN{Helen Yannakoudakis}
\IEEEauthorblockA{
King's College London\\
London, UK}
\and
\IEEEauthorblockN{Jie M. Zhang}
\IEEEauthorblockA{
King's College London\\
London, UK}
}

\fi


%


\maketitle

\ifreview
\else
    \thispagestyle{fancy}
\fi

\begin{abstract}
Explicit reasoning can help language models generate code, but it also introduces a practical risk: the model may spend so much of its output budget reasoning that it never produces usable code.
We study this behaviour in relatively small open-weight reasoning models, framing it as \emph{reasoning-trace overflow}.
Across seven models and six code benchmark settings, we measure how much of a fixed 32k-token output budget is spent on reasoning, whether that reasoning improves \textit{pass@1}, and whether lightweight inference-time controls can mitigate overflow.
Our results show that the value of reasoning is highly conditional.
Overflow is rare on simpler function-level tasks, but becomes common on harder code-generation benchmarks and for particular model families: on several model--benchmark pairs, more than half of generations exhaust the full 32k-token budget while still reasoning.
Most complete overflow is associated with repetitive reasoning loops, although substantial long reasoning also occurs without looping.
Disabling reasoning can therefore be competitive for overflow-prone models, simply because it ensures that an answer is produced.
At the same time, reasoning remains valuable for stronger models on harder algorithmic and code-understanding tasks.
Finally, prompt steering can substantially reduce overflow for some models but is inconsistent, while budgeted transitions provide more direct control and can improve both performance and token efficiency for overflow-prone models.
These findings suggest that explicit reasoning should be treated as a controllable resource for code generation, rather than an always-on capability.
\end{abstract}


%
\IEEEpeerreviewmaketitle

\section{Introduction}

Relatively small open-weight reasoning models are becoming increasingly attractive for code generation.
They can be run locally, provide greater control over inference, and expose explicit reasoning traces that can be inspected and manipulated~\cite{wangShortSurveySmall2026}.
Recent work has shown that reasoning can support code generation through structured prompting, planning, and lightweight model training~\cite{liStructuredChainofThoughtPrompting2025,jiangSelfPlanningCodeGeneration2024,yangChainofThoughtNeuralCode2024}, while survey work highlights code reasoning as an increasingly important direction for software engineering~\cite{pujarCodeReasoningSoftware2026}.
However, explicit reasoning is not free: longer traces consume additional tokens and inference compute, and can contain substantial unnecessary reasoning~\cite{wuThoughtCalibrationEfficient2025,hanTokenBudgetAwareLLMReasoning2025a}.

This raises a practical question: when models are asked to generate code, is explicit reasoning actually worth the cost?
For code generation, this creates an additional constraint.
A model must not only reason about the task, but also leave enough output budget to produce a complete program.
For harder tasks, these two demands can grow together: the model may spend more tokens planning, searching, or checking possible implementations, while the final answer itself may also require more tokens.
When reasoning consumes the available output budget, the model can fail slowly, spending substantial inference time and compute without returning usable code to the developer.

We refer to this failure mode as \emph{reasoning-trace overflow}.
Unlike ordinary verbosity, reasoning-trace overflow prevents answer generation altogether.
Recent work has explored controlling reasoning length through token budgets, early stopping, and explicit reasoning limits~\cite{hanTokenBudgetAwareLLMReasoning2025a,muennighoffS1SimpleTesttime2025a,liaoFracturedChainofThoughtReasoning2025,xuScalableChainThoughts2025a}.
Most of this work focuses on controlling reasoning length in mathematical and general reasoning settings, with comparatively little attention paid to how excessive reasoning naturally manifests during code generation.
This setting introduces an additional constraint: the reasoning trace must share the output budget with a potentially long executable answer.

To fill this gap, we conduct an empirical study of \textit{reasoning-trace overflow} in code generation.
We evaluate seven open-weight reasoning models across six code benchmark settings under a fixed $32k$-token output budget.
Our study asks three questions.
First, how much of the output budget do these models spend on reasoning, and when do they overflow?
Second, does extended reasoning improve accuracy enough to justify this cost?
Third, can lightweight inference-time controls -- either prompt-based steering, which influences reasoning behaviour without modifying decoding, or budgeted transitions, which directly control when reasoning ends -- reduce overflow while preserving accuracy?

Our results show that the value of explicit reasoning is highly conditional.
\textit{Reasoning-trace overflow} is uncommon on simple function-level synthesis tasks, but becomes frequent on harder code-generation benchmarks and for particular model families.
Most complete overflow is associated with repetitive reasoning loops, although substantial long reasoning also occurs without looping.
When overflow is common, disabling reasoning can be competitive not because reasoning is necessarily unhelpful, but because it ensures that the model reaches the answer stage: a response that exhausts its budget while still reasoning has no opportunity to pass the tests.
At the same time, reasoning remains valuable for larger models on harder algorithmic and code-understanding benchmarks, where it can substantially improve \textit{pass@1} when the model successfully transitions to an answer.

We also find that overflow can be mitigated at inference time, but not reliably across all settings.
Prompt steering can substantially reduce overflow for some models, but its effects are inconsistent and can come at the cost of accuracy.
Budgeted transitions provide more direct control and can improve performance for overflow-prone models, while substantially reducing the token cost of successful answers.
Overall, our findings suggest that reasoning should be treated as a controllable resource for code generation, rather than an always-on capability.

\smallskip
\noindent\textit{Our contributions are as follows:}

\begin{enumerate}[leftmargin=*]
    \item We define and operationalise \emph{reasoning-trace overflow}, a failure mode in which explicit reasoning consumes the output budget before a code solution is produced.

    \item We provide a broad empirical characterisation of reasoning-trace overflow across seven relatively small open-weight reasoning models and six code benchmark settings, showing how it varies with task type, model family, and model scale, how often complete overflow coincides with repetitive reasoning loops, and whether the associated reasoning improves accuracy.
    
    \item We compare complementary forms of inference-time reasoning control, from prompt-based soft constraints to enforced transition budgets, and show how they trade off overflow, accuracy, and token efficiency.

\end{enumerate}


\section{Related Work}

\paragraph{\textbf{Reasoning for Code Generation}}
Chain-of-thought (CoT) prompting showed that intermediate reasoning can improve performance on complex tasks~\cite{weiChainofthoughtPromptingElicits2022}.
Subsequent work has explored more structured forms of reasoning, including Program-of-Thoughts prompting, which expresses reasoning through executable programs~\cite{chenProgramThoughtsPrompting2023b}, and Structured CoT prompting, which guides code generation using program structures such as sequences, branches, and loops~\cite{liStructuredChainofThoughtPrompting2025}.
For code generation specifically, Self-Planning separates high-level planning from implementation~\cite{jiangSelfPlanningCodeGeneration2024}, while COTTON studies CoT reasoning for lightweight code-generation models~\cite{yangChainofThoughtNeuralCode2024}.
More recent work has also explored selective reasoning for code generation, such as UnCert-CoT, which activates CoT reasoning based on uncertainty to avoid unnecessary overthinking~\cite{zhuUnCertCoTUncertaintyAwareChainofThought2025}.
A recent survey highlights that reasoning is increasingly central to software engineering tasks, while also identifying the behaviour and failure modes of code reasoning as an open area for further study~\cite{pujarCodeReasoningSoftware2026}.
Our work follows this direction by studying not only whether reasoning helps code generation, but also when it consumes the output budget needed to produce code.

\paragraph{\textbf{Long Reasoning}}
Recent work increasingly treats reasoning as a test-time resource rather than a free capability.
Longer reasoning can produce diminishing or even negative returns~\cite{wuWhenMoreLess2025a}, motivating efficient-reasoning methods that reduce unnecessary thinking while preserving accuracy~\cite{suiStopOverthinkingSurvey2025b,hanTokenBudgetAwareLLMReasoning2025a}.
This broader phenomenon is often described as \textit{overthinking}, where models continue reasoning after useful returns diminish or produce redundant computation~\cite{zhangLLMsReallyNeed2026,zhouWhenMoreThinking2026}.
Recent work has also identified repetitive reasoning loops as a distinct failure mode in reasoning models~\cite{pipisWaitWaitWait2026,duanCircularReasoningUnderstanding2026}.
However, these studies do not consider code generation, where the reasoning trace must also leave sufficient output budget for an executable answer.
Our focus is therefore a distinct operational failure: the reasoning trace can exhaust this shared budget before the model produces any usable code.

\paragraph{\textbf{Reasoning Control}}
Several recent approaches control reasoning length more directly.
S1 introduces budget forcing to terminate reasoning~\cite{muennighoffS1SimpleTesttime2025a}, Fractured CoT studies answering from truncated reasoning traces~\cite{liaoFracturedChainofThoughtReasoning2025}, Elastic Reasoning separates thinking and solution budgets~\cite{xuScalableChainThoughts2025a}, and TRACES explores adaptive early stopping~\cite{belkhiterTRACESTaggingReasoning2026}.
In code specifically, Z1 combines code-reasoning training with a shifted thinking window to encourage answers~\cite{yuZ1EfficientTesttime2025}.
We do not claim these budget-control mechanisms as new.
Instead, we study the behaviour that motivates them in code generation: when existing reasoning models naturally exhaust their output budget before producing an answer, whether this extended reasoning improves correctness, and how soft and hard controls trade off overflow, accuracy, and token efficiency.
We additionally measure the token cost per correct answer, capturing whether reasoning control produces useful code more efficiently rather than simply shortening the reasoning trace.


\section{Reasoning-Trace Overflow}

Explicit reasoning models produce a reasoning trace before emitting a final answer.
In code generation, this can be costly because the model must both reason about the implementation and still leave enough output budget to produce the code itself.
Longer or more complex solutions may therefore create pressure on both sides: they can require more reasoning to plan, and more tokens to write.

We define \textbf{reasoning-trace overflow} as a failure mode in which the explicit reasoning trace consumes the available output budget, preventing the model from producing a solution.

Formally, for a given input and decoding run, let \(B\) denote the operational output budget, and let \(y\) be the generated output token sequence.
We decompose \(y\) into a reasoning trace \(y_r\) and an answer \(y_a\), such that \(y = y_r y_a\), where either segment may be empty depending on the generation behaviour.

We say that \textbf{\textit{reasoning-trace overflow}} occurs when generation terminates at the output budget while still inside the reasoning trace:
\[
|y| = B \;\land\; |y_r| = B \;\land\; |y_a| = 0.
\]

This is distinct from reasoning that is simply long or verbose; extended reasoning may still be useful when it leads to better code.
The failure occurs when the reasoning trace crowds out the solution itself, causing the model to spend time and compute \textit{thinking} without producing executable code.

We use reasoning-trace overflow as the central lens for our study.
We first measure when explicit reasoning models approach or exhaust a fixed output budget, then examine whether extended reasoning improves accuracy enough to justify its cost, and finally evaluate whether lightweight inference-time controls can mitigate overflow.


\pagebreak

\section{Research Questions}

\begin{rqlist}[left=0pt]

  \item
  \textbf{Overflow Dynamics.}
  \textit{When do reasoning models approach or exhaust their operational token budget on code-generation tasks?}
    We characterise reasoning overflow across models and benchmarks, analysing how reasoning-token usage varies across task types and benchmark difficulty profiles, including short-answer tasks that help separate reasoning cost from answer length.
    We further examine whether complete overflow is associated with repetitive reasoning loops.

  \item
  \textbf{Reasoning Utility.}
  \textit{Does extended reasoning improve coding task accuracy enough to justify its cost?}
  We compare reasoning-enabled and reasoning-disabled generation to analyse when reasoning improves correctness, when it has little effect, and when it harms performance.

  
  \item
  \textbf{Reasoning Control.}
  \textit{Can lightweight inference-time controls reduce reasoning-trace overflow while preserving coding task accuracy?}
  We evaluate both prompt-based steering and budgeted reasoning-to-answer transitions, examining whether these interventions provide reliable control across models and tasks.

\end{rqlist}


\section{Experimental Set-up}

We conduct a controlled benchmarking study of explicit reasoning behaviour across seven models and six code benchmark settings.
Our repository provides the complete implementation and reproducibility details, including exact model and dataset revisions, software and library versions, generated outputs, per-task results, and evaluation configurations.\footnote{\repourl}

\subsection{Model Selection}

We focus on relatively small open-weight models because they are practical candidates for private, low-cost code generation, while providing full access to and control over their reasoning traces~\cite{wangShortSurveySmall2026}.
We evaluate seven open-weight explicit reasoning models, summarised in Table~\ref{tab:models}.
All models expose an explicit reasoning trace before producing a final answer, allowing us to separate reasoning from answer tokens and directly measure reasoning length and its effect on accuracy.

Our selection is designed to cover three sources of variation.
First, we include models from different families and providers, including Alibaba's \texttt{Qwen} models~\cite{yangQwen3TechnicalReport2025}, the distillation-driven \llama~\cite{guoDeepSeekR1IncentivizesReasoning2025}, the fully open \olmo~\cite{olmoOlmo32026}, and \texttt{NVIDIA}'s Nemotron models~\cite{ahmadOpenCodeReasoningAdvancingData2025a}.
This reduces the risk of drawing conclusions from a single architecture or post-training recipe.
Second, the three \texttt{Qwen} models provide a controlled size comparison from 8B to 32B parameters, allowing us to study how \textit{reasoning-trace overflow} changes with model scale.
Third, the two Nemotron models provide a useful specialisation comparison: \openr\ targets general reasoning, while \opencr\ is specialised for code generation.
This lets us examine whether code-specific reasoning training reduces overflow or instead changes how reasoning budget is allocated.

\subsection{Model Configuration}

We run all models with a fixed output budget of $32k$ tokens, following prior work that evaluates reasoning models under large but finite generation budgets~\cite{liaoFracturedChainofThoughtReasoning2025}.
Although some models support longer outputs, $32k$ tokens is already a substantial allowance for interactive code generation without additional repository context, and is commonly used as a large-budget setting for reasoning-model evaluation~\cite{QuickstartQwen,EvaluationReasoningModels}.
We deliberately use this generous budget to avoid inducing overflow through an artificially tight output limit; smaller operational budgets would only expose additional generations that had not yet completed reasoning.
Using the same output budget for every model also provides a shared resource constraint for comparison.
Therefore, our reported overflow rates are operational measurements under a $32k$-token budget.

We use greedy decoding throughout ($temperature=0.0$, $top\_p=1.0$) as a controlled, low-variance setting commonly used in code-generation evaluation~\cite{songGoodBadGreedy2025}.
Each model therefore produces a single solution per task, which we evaluate using \textit{pass@1}.
Generation terminates when the model produces its end-of-sequence token or reaches the $32k$-token output budget.
We use each model's native chat template, preserving its expected reasoning and answer format, and do not use a system prompt so that models are evaluated under the task prompt alone without additional steering~\cite{muCloserLookSystem2025}.
All experiments are run directly on a single NVIDIA H200 GPU.

\begin{table}[t]
\centering
\caption{Details of all reasoning models evaluated in our study.}
\label{tab:models}
\small
\setlength{\tabcolsep}{2.5pt}

  \begin{adjustbox}{width=\columnwidth}
  \bgroup
  \def\arraystretch{1.5}
\begin{tabular}{llllll}
\toprule
\textbf{Name} & \textbf{Provider} & \textbf{Variant} &
\textbf{Size} & \textbf{Release} & \textbf{Licence} \\
\midrule
\rowcolor{summarybg}
\qwlil  & Qwen     & Qwen3~\cite{QwenQwen38BHugging2025a}
        & 8B  & Apr.\ 2025 & Apache 2.0 \\
\qwmid  & Qwen     & Qwen3~\cite{QwenQwen314BHugging2025}
        & 14B & Apr.\ 2025 & Apache 2.0 \\
\rowcolor{summarybg}
\qwbig  & Qwen     & Qwen3~\cite{QwenQwen332BHugging2025}
        & 32B & Apr.\ 2025 & Apache 2.0 \\
\llama  & DeepSeek & R1 Distilled~\cite{DeepseekaiDeepSeekR1DistillLlama8BHugging2025}
        & 8B  & Jan.\ 2025 & MIT \\
\rowcolor{summarybg}
\olmo   & Ai2      & OLMo 3 Think~\cite{AllenaiOlmo37BThinkHugging2026a}
        & 7B  & Dec.\ 2025 & Apache 2.0 \\
\openr  & NVIDIA   & OpenReasoning~\cite{NvidiaOpenReasoningNemotron7BHugging2026a}
        & 7B  & Jul.\ 2025 & CC BY 4.0 \\
\rowcolor{summarybg}
\opencr & NVIDIA   & OpenCodeReasoning 1.1~\cite{NvidiaOpenCodeReasoningNemotron117BHugging2025}
        & 7B  & Jul.\ 2025 & NVIDIA OML \\
\bottomrule
\end{tabular}
  \egroup
  \end{adjustbox}
\end{table}

\subsection{Dataset Selection}

We select established benchmarks that span distinct coding task styles while providing automated correctness evaluation, covering function-level synthesis, practical library-based programming, competitive programming, and short-answer code reasoning.
We include \evalplus~\cite{liuYourCodeGenerated2023} as a clean function-level synthesis setting, using its \texttt{HumanEval+}~\cite{UploadREADMEmdHuggingface_hub2024} and \texttt{MBPP+}~\cite{RemoveTestjsonlUse2024} tasks (542 tasks).
We include \bigcodebench~\cite{zhuoBigCodeBenchBenchmarkingCode2024,UpdateREADMEmdBigcode2025} to test practical \python\ tasks with complex instructions and diverse library/function calls (1,140 tasks).
We use \livecodebench~\cite{jainLiveCodeBenchHolisticContamination2024,FixTypos42025} to evaluate recent and contamination-aware competitive-programming problems, using the \texttt{v5} and \texttt{v6} subsets (342 tasks).
Finally, we use the \texttt{test} subset of \codecontests~\cite{doi:10.1126/science.abq1158,MinorREADMEImprovement2023} as an additional competitive-programming benchmark with challenging algorithmic problems (165 tasks).

\begin{figure*}[t]
    \centering
    \includegraphics[
    width=\textwidth,
    alt={
Six panels compare reasoning-token usage across seven models and
six code benchmark settings. Overflow is generally low on simpler
synthesis tasks, but rises substantially on LiveCodeBench and
CodeContests. CRUX-Input also produces high overflow for several
Llama and NVIDIA models despite requiring short answers. Within the
Qwen family, reasoning-token usage and overflow generally decrease
as model size increases from 8B to 32B.
}
]{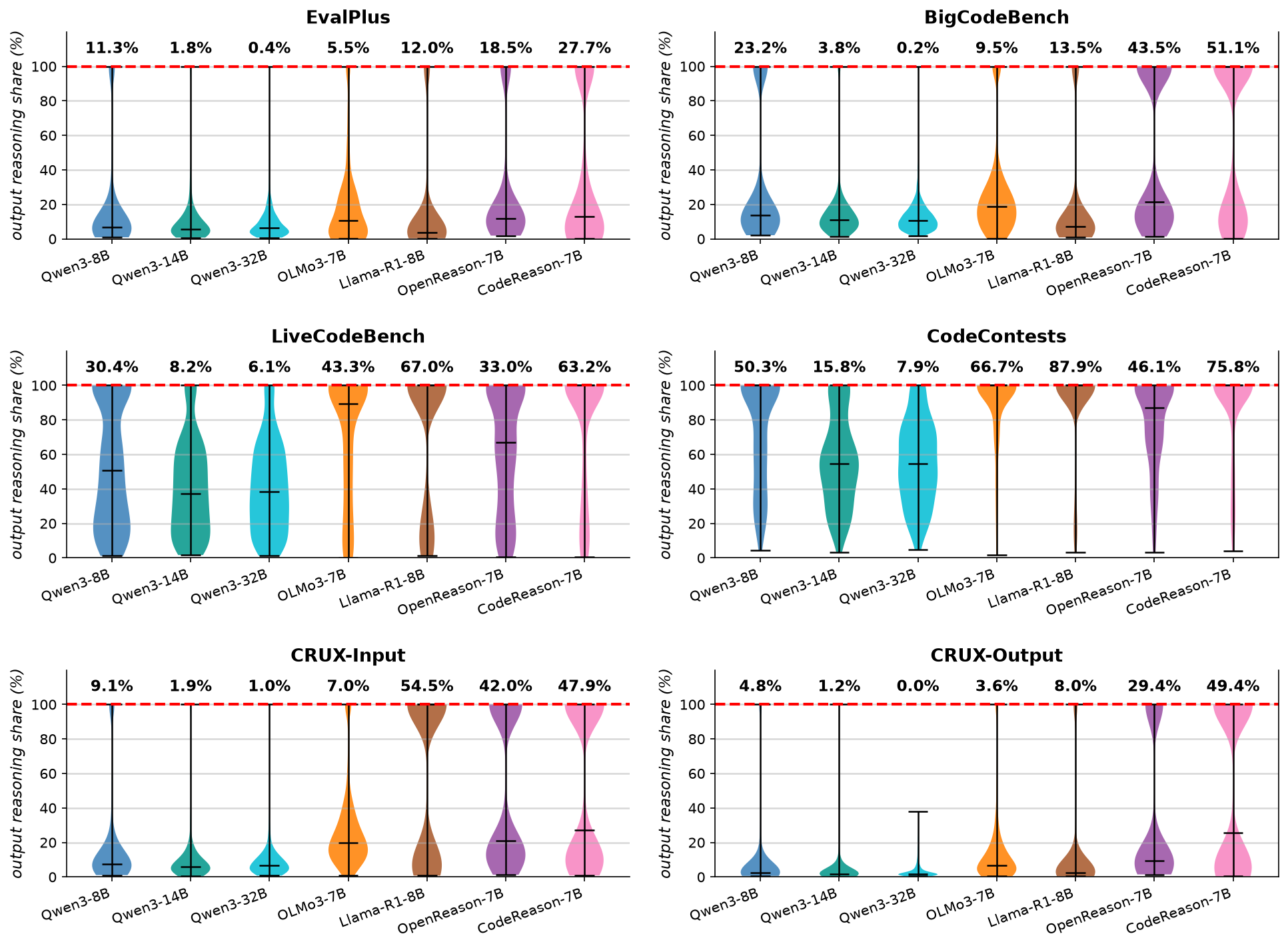}
    \caption{
        \textbf{\textit{Results for RQ1 Overflow Dynamics.}}
        Distribution of reasoning-token usage under the standard reasoning-enabled setting.
        Each violin shows the percentage of the 32k output budget consumed by the explicit reasoning trace for a given model and benchmark.
        The dashed red line marks the full output budget, and the percentage above each violin reports the \textit{reasoning-trace overflow} rate.
    }
    \label{fig:overflow}

\end{figure*}

These benchmarks let us study reasoning-trace overflow across tasks that vary in difficulty profile, output length, and required programming knowledge.
However, code generation also conflates reasoning with answer production, since successful answers often require many output tokens.
To separate these factors, we additionally evaluate on \crux~\cite{guCRUXEvalBenchmarkCode2024,AddMetadataDataset2024}, a code-reasoning benchmark built from short \python\ functions with input-prediction (\cruxin) and output-prediction (\cruxout) tasks (800 tasks each).
\crux\ allows us to test whether reasoning traces become long for code-based tasks even when the final answer is short.

\subsection{Evaluation Details}

For each benchmark, we use the evaluation pipeline and execution environment recommended by the original benchmark authors, ensuring that \textit{pass@1} is computed consistently with the benchmark's task formatting, tests, and correctness checks.

All models in our study expose reasoning using the standard \texttt{<think>} and \texttt{</think>} delimiters.
We use \texttt{ThinkPack}~\cite{twistReasoningCollapse2026} to parse each output into a reasoning trace and an answer segment using these tags.
For reasoning-enabled inference, the reasoning trace is the content generated before \texttt{</think>}, and the answer is the content generated after.
For reasoning-disabled inference, we insert an empty reasoning segment before generation, forcing the model to begin directly in the answer.

We count reasoning tokens using each model's own tokenizer.
A run is counted as \textit{reasoning-trace overflow} when generation reaches the $32k$ output budget without having produced the \texttt{</think>} tag, leaving no answer segment.
In overflow cases, we do not attempt to extract code from the unfinished reasoning trace.
Reasoning traces can be difficult or even illegible to users~\cite{joseReasoningModelsSometimes2026}, and without the closing delimiter the model has not transitioned to its answer under the expected output protocol.

We additionally detect sustained textual reasoning loops across all reasoning-enabled generations.
For each trace, we match its final 64-character suffix to the nearest earlier occurrence, using the distance between them as a candidate repetition period.
We then trace this periodic pattern backwards until it breaks, and flag a loop only when the repeated region spans at least 2,000 characters and contains at least three complete cycles.
These conservative thresholds avoid treating short or incidental repetition as a loop.

All reported metrics are computed from the stored per-task generations using the analysis scripts released in our repository.\footnote{\repourl}


\section{Experimental Results}


\subsection{RQ1 Overflow Dynamics}

\paragraph{\textbf{Set-up}}

To measure reasoning-trace overflow, we run each model on each benchmark using the standard task prompt and a fixed $32k$ output budget.
For each response, we detect whether \textit{reasoning-trace overflow} occurred and calculate the percentage of the output budget used by the reasoning trace.
\textit{Fig.~\ref{fig:overflow} shows the resulting distributions.}

\paragraph{\textbf{Results}}

Reasoning-trace usage varies substantially across both models and benchmarks.
On \evalplus, overflow is generally rare, with most models using only a small fraction of the available budget for reasoning.
This is expected, as \evalplus\ contains relatively small function-level synthesis tasks.
However, even here, behaviour is not uniform: \openr\ overflows on 27.7\% of tasks, while the larger \texttt{Qwen} models almost never overflow.

The harder generation benchmarks show a much clearer overflow problem.
On \livecodebench\ and \codecontests, several models spend a large fraction of the $32k$ budget on reasoning, and overflow rates become substantial.
This suggests that reasoning-trace overflow is strongly associated with harder code-generation settings, where models have more to reason about while still needing to reserve enough tokens for a complete solution.
The contrast with \bigcodebench\ is informative.
Although \bigcodebench\ includes practical tasks involving libraries and APIs, its overflow rates are generally lower than those on \livecodebench\ and \codecontests.
This suggests that inherent problem difficulty may be a stronger driver of overflow than API or library use alone.
Models appear to overflow not simply because code answers are long, but because some tasks invite extended search, planning, or exploration before code generation.

The \crux\ results provide a useful contrast.
Because \crux\ requires short answers rather than full code generation, it helps separate reasoning cost from answer length.
Most models show low overflow on \cruxout, suggesting that short-answer code reasoning is less likely to exhaust the budget.
However, \cruxin\ produces high overflow for several models, including \llama, \openr, and \opencr.
This shows that overflow is not only caused by long code outputs: some code-based tasks can trigger long reasoning traces even when the final answer is short.

Finally, the \texttt{Qwen} family shows a consistent size effect.
Across all benchmarks, overflow decreases sharply as model size increases from \qwlil\ to \qwmid\ and \qwbig.
This suggests that larger models may require less extended reasoning to reach a solution, or may be better calibrated in when to stop reasoning.
This is particularly visible on \evalplus, \bigcodebench, and \crux, where \qwbig\ almost never overflows.

\begin{figure}[t]
    \centering
    \includegraphics[
    width=\columnwidth,
    alt={
        Bar chart showing the percentage of overflowed generations
        containing a detected reasoning loop for seven models.
        Loop prevalence is 98\% for Qwen3-8B, 98\% for Qwen3-14B,
        85\% for Qwen3-32B, 62\% for OLMo3-7B, 100\% for
        Llama-R1-8B, 94\% for OpenReason-7B, and 96\% for
        CodeReason-7B.
    }
]{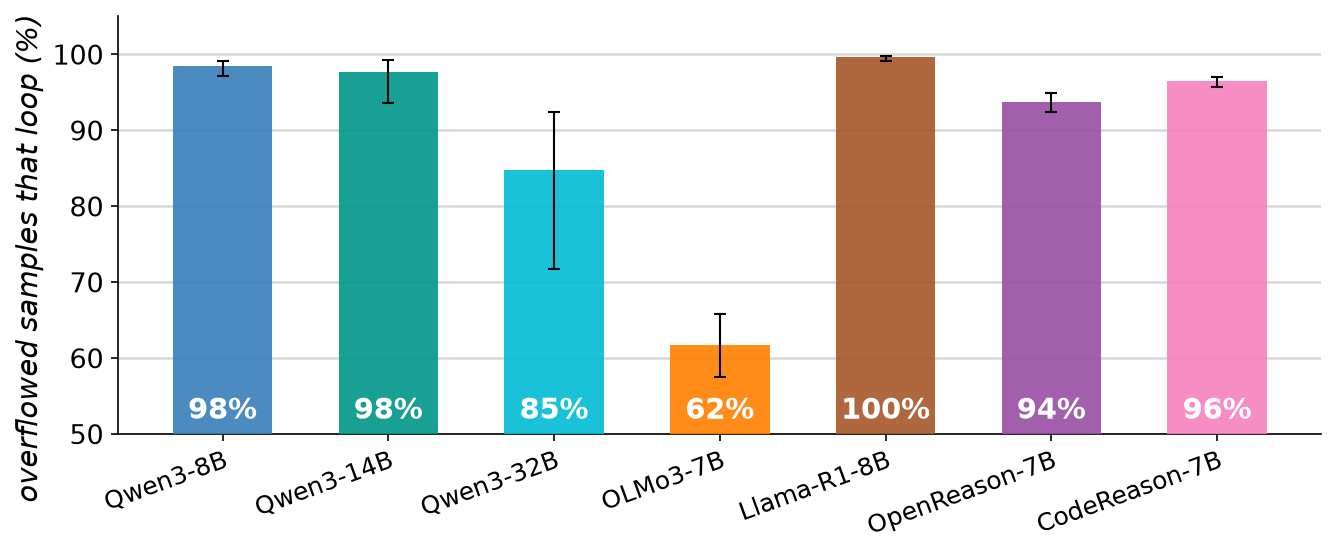}
    \caption{
        \textbf{\textit{Reasoning Loop Analysis.}}
        Percentage of overflowed generations containing a detected
        reasoning loop for each model.
        A loop is identified as a repeated region spanning at least
        2,000 characters and three complete cycles.
    }
    \label{fig:loops}
\end{figure}

\paragraph{\textbf{Reasoning Loops}}
Reasoning loops are strongly associated with \textit{reasoning trace overflow}.
Overall, 94\% of overflowed generations contain a detected loop.
As shown in Fig.~\ref{fig:loops}, this is consistent across most models, ranging from 85--100\%, with \olmo\ lower at 62\%.

However, long reasoning is not limited to looping behaviour.
Among generations using more than half of the 32k reasoning budget, 22\% contain no detected loop.
More strikingly, 99\% of generations that reason for more than 16k tokens but still produce an answer contain no detected loop.
Together with the broad reasoning-length distributions in Fig.~\ref{fig:overflow}, this suggests that loops are a major mechanism of complete overflow, while extended non-looping reasoning remains common on harder coding tasks.

\smallskip
\begin{summarybox}{RQ1 Summary.}
Reasoning-trace overflow is not a rare edge case, but it is highly task- and model-dependent.
It is uncommon on simple function-level synthesis, much more common on harder code-generation benchmarks, and still possible on short-answer code-reasoning tasks.
Most complete overflow is associated with detected reasoning loops, although substantial long reasoning also occurs without looping.
These results motivate the next question: whether extended reasoning improves accuracy enough to justify this risk.
\end{summarybox}


\begin{figure*}[t]
    \centering
    \includegraphics[
        width=\textwidth,
        alt={
Grouped bars compare pass@1 with and without explicit reasoning
across seven models and six benchmark settings. Explicit reasoning
provides the largest gains for the Qwen models on harder generation
and code-reasoning tasks, while reasoning-disabled inference is often
competitive on simpler generation tasks. OLMo and the NVIDIA models
show more mixed effects, demonstrating strong model- and task-dependent
variation in the utility of reasoning. 
}
]{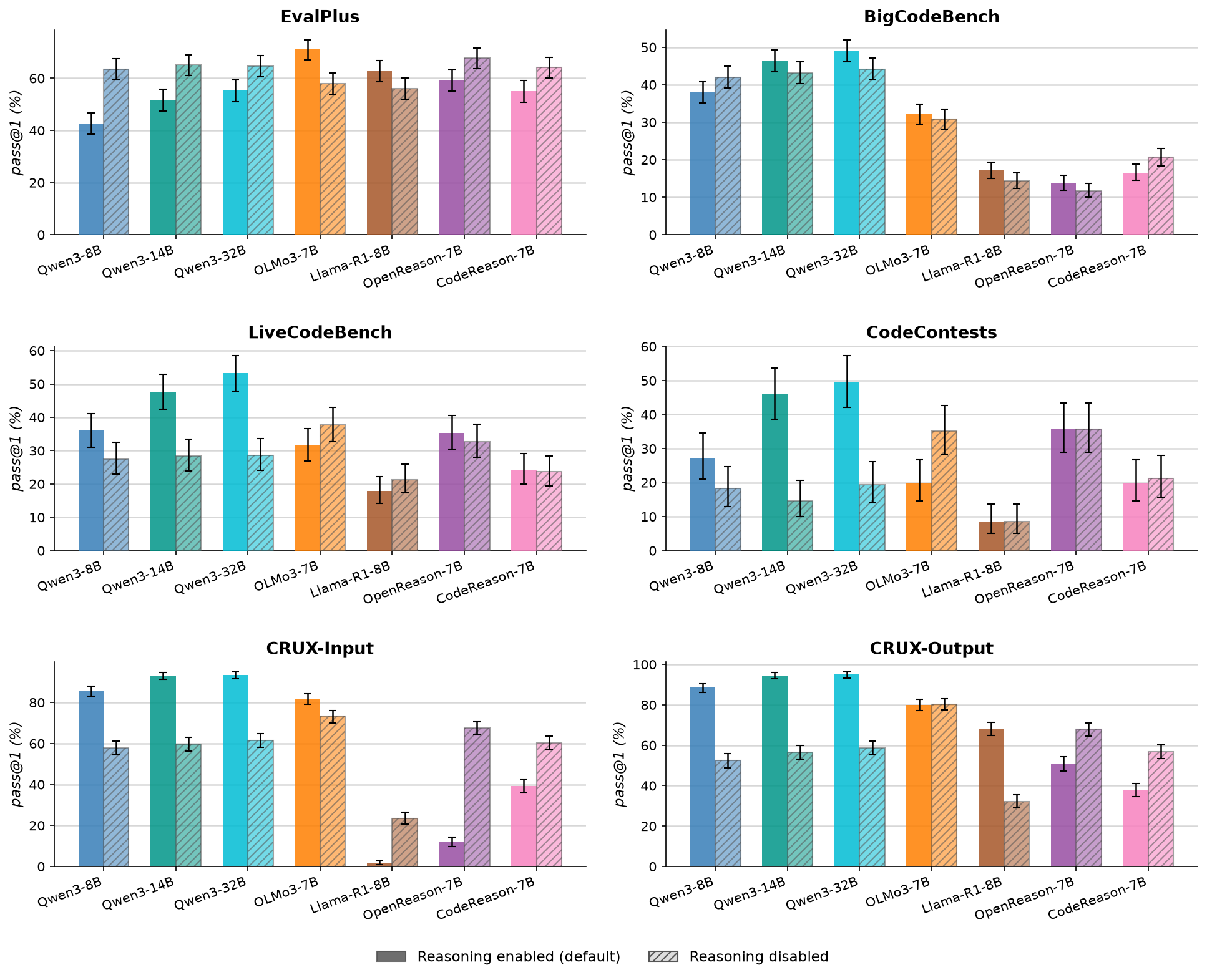}
    \caption{
        \textbf{\textit{Results for RQ2 Reasoning Utility.}}
        Model accuracy (\textit{pass@1}) with and without reasoning across all models and benchmarks.
        Solid bars show the standard reasoning-enabled setting, while hatched bars show the reasoning-disabled setting, where empty reasoning tags are inserted before generation to force the model to answer directly.
        Error bars indicate 95\% Wilson confidence intervals.
    }
    \label{fig:accuracy}
    
\end{figure*}

\subsection{RQ2 Reasoning Utility}

\paragraph{\textbf{Set-up}}

To evaluate whether extended reasoning is useful, we compare standard reasoning-enabled inference against a reasoning-disabled setting.
Here, reasoning-disabled refers specifically to disabling the explicit reasoning trace in the response; it does not imply the absence of latent internal computation, and the final answer may still contain reasoning-like text.
This comparison is especially relevant given RQ1, where reasoning-enabled inference can fail to produce an answer when the reasoning trace exhausts the output budget.
\textit{Accuracy comparisons are given in Fig.~\ref{fig:accuracy}.}

\begin{figure*}[t]
    \centering
    \includegraphics[
    width=\textwidth,
    alt={
Two grouped bar charts compare macro-average pass@1 and
reasoning-trace overflow across five prompt styles for seven models.
Prompt effects vary substantially by model. The aspects prompt sharply
reduces Qwen3-8B overflow with little accuracy loss, while OLMo also
benefits from brevity-oriented prompts. OpenReason-7B overflow worsens
under every steering prompt, and qualitative brevity prompts substantially
reduce Qwen3-32B accuracy. The explicit token-budget prompt provides
only weak control for most models.
}
]{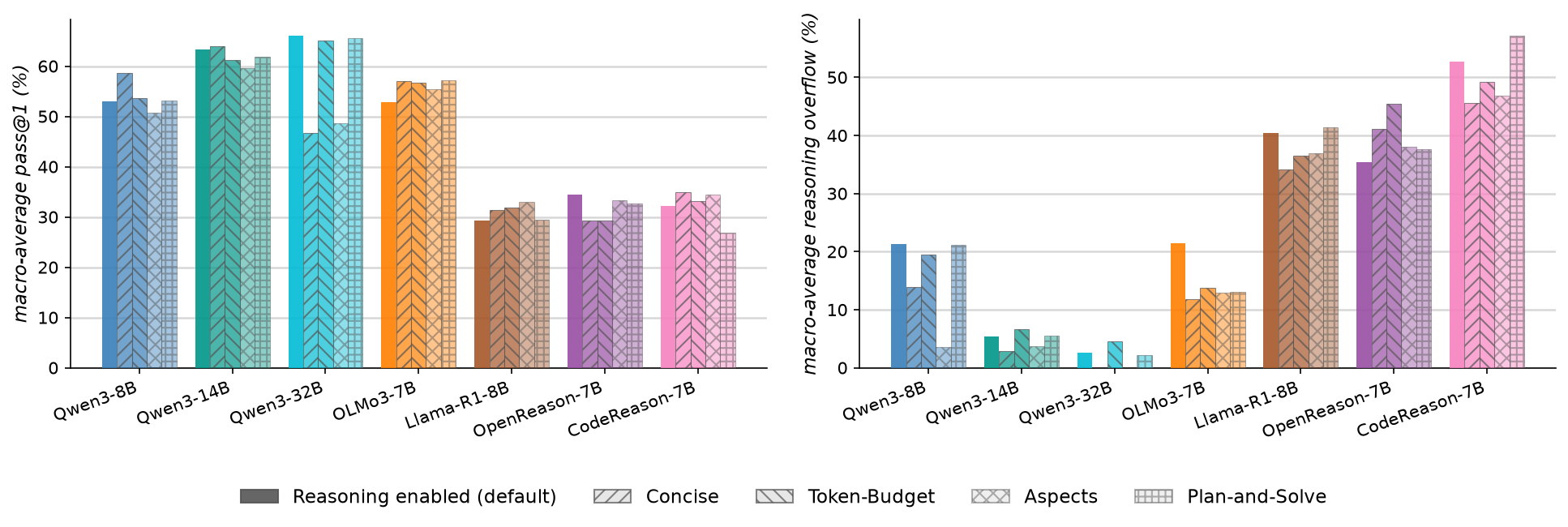}
    \caption{
        \textbf{\textit{Results for RQ3 Reasoning Control, Prompt Steering.}}
        Effect of prompt-based steering on reasoning behaviour.
        Bars show macro-average \textit{pass@1} and \textit{reasoning-trace overflow} across all benchmark settings for each model under the reasoning enabled, concise, token-budget, aspects, and Plan-and-Solve prompt-based strategies.
    }
    \label{fig:steering}

\end{figure*}

\paragraph{\textbf{Results}}

Extended reasoning is not uniformly beneficial.
On \evalplus\ and \bigcodebench, reasoning-disabled inference is often competitive with reasoning-enabled inference, with no consistent best option across models.
This is important because reasoning-enabled inference can fail to produce an answer due to reasoning-trace overflow, wasting many tokens, whereas reasoning-disabled inference always proceeds directly to an answer.
For these benchmarks, the additional cost and risk of explicit reasoning is not consistently matched by higher accuracy.

The harder generation benchmarks show a clearer benefit for some models, but still not universally.
For the \texttt{Qwen} models, reasoning substantially improves performance on \livecodebench\ and \codecontests.
For example, when reasoning is enabled, \qwbig\ improves by 25 and 30 percentage points on \livecodebench\ and \codecontests, respectively.
This aligns with the intuition that harder algorithmic tasks can benefit from explicit reasoning.
However, the pattern is different for models that showed higher reasoning-trace overflow on these benchmarks.
For \olmo, reasoning-disabled inference performs similarly or better on several benchmarks, while \llama\ remains weak on the harder generation tasks regardless of reasoning mode.
The \texttt{NVIDIA} models also show mixed behaviour: \opencr\ benefits slightly from reasoning on \codecontests, while \openr\ changes little.
This suggests that the utility of reasoning is strongly model- and task-dependent.
When a model is likely to exhaust its budget while reasoning, the benefit of explicit reasoning may be outweighed by the risk of never producing an answer.

The \crux\ results show that short-answer code reasoning behaves differently from code generation.
For the \texttt{Qwen} models, explicit reasoning is especially useful: across \cruxin\ and \cruxout, reasoning-enabled inference is on average 34 percentage points higher than reasoning-disabled inference.
This indicates that these code-understanding tasks benefit from reasoning, even though their final answers are short.
In contrast, the \texttt{NVIDIA} models often improve when reasoning is disabled, despite showing high overflow on \cruxin\ in RQ1.
This suggests that some models may over-reason on short-answer tasks, spending substantial budget without gaining accuracy.

\smallskip
\begin{summarybox}{RQ2 Summary.}
Extended reasoning helps in some settings, especially for stronger models on harder algorithmic and code-reasoning tasks, but only when the model is able to produce a final answer.
On simpler code-generation tasks, reasoning can add substantial cost and occasionally prevent answer generation without delivering consistent accuracy gains.
A budget-exhausted reasoning trace has no opportunity to pass the tests, regardless of whether its partial reasoning was useful.
Therefore, when overflow is likely, disabling reasoning may improve accuracy simply by ensuring that the model produces an answer.
\end{summarybox}

\begin{figure*}[t]
    \centering
    \includegraphics[
        width=\textwidth,
        alt={
Lines show macro-average pass@1 for seven models as the enforced
reasoning-transition budget increases from immediate transition to
28k tokens. Most models improve rapidly after being allowed a small
amount of reasoning and then begin to plateau. CodeReason-7B gains
substantially from a moderate transition budget, while OpenReason-7B
performs best with immediate transition. The best reasoning budget
varies across models.
}
]{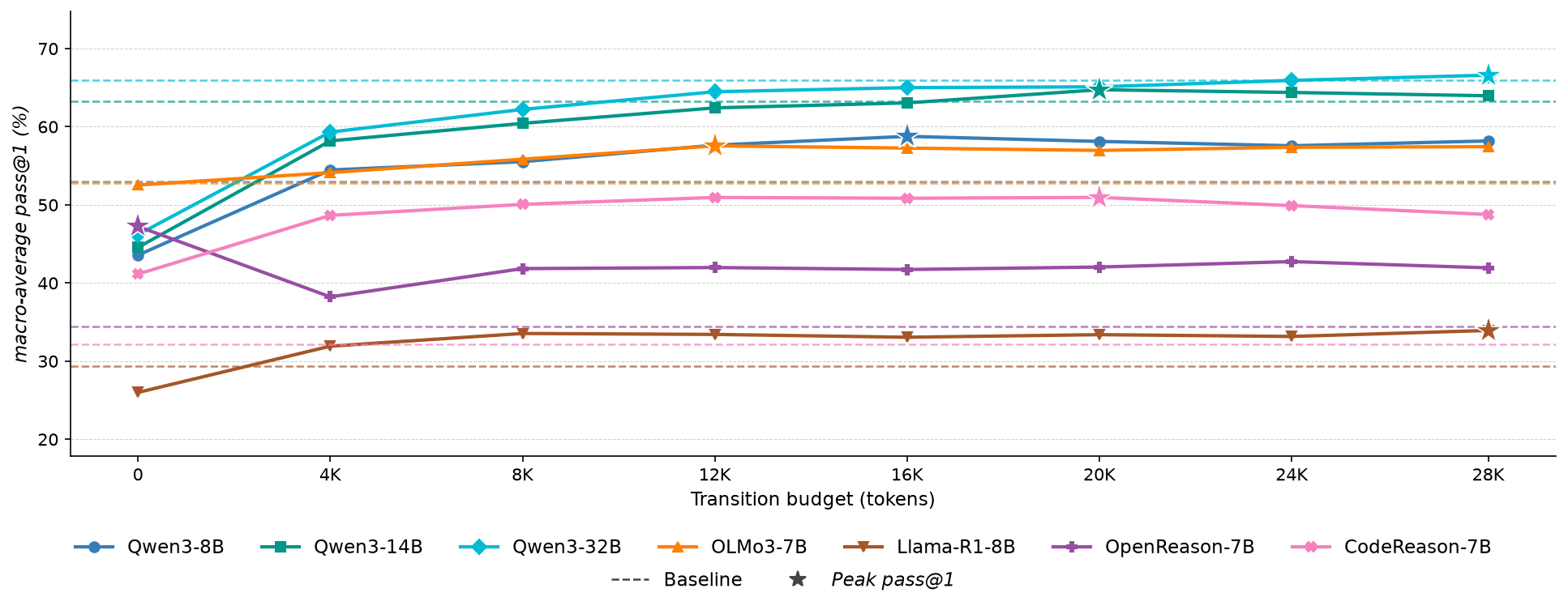}
    \caption{
        \textbf{\textit{Results for RQ3 Reasoning Control, Budgeted Transitions.}}
        Effect of enforced reasoning-transition budgets on accuracy.
        Each line shows macro-average \textit{pass@1} across all benchmark settings as the transition budget varies from immediate transition to 28k tokens.
        Dashed lines show each model's standard reasoning-enabled baseline, and stars mark the best observed transition budget.
    }
    \label{fig:transition}

\end{figure*}

\subsection{RQ3 Reasoning Control}

We investigate two complementary inference-time approaches to controlling reasoning. Prompt steering provides soft control by influencing reasoning behaviour without modifying decoding, while budgeted transitions provide hard control by explicitly enforcing when reasoning ends.
For space, we report macro-average results in the main paper, weighting each of the six benchmark settings equally; exact prompts and full results are available in our repository.\footnote{\repourl}

\subsubsection{\textbf{Prompt Steering}}

\paragraph{\textbf{Set-up}}

We first test whether reasoning can be controlled through prompt steering alone, by appending a fixed instruction to the original task prompt.
We compare five prompt styles:
an unmodified \textit{baseline} (the default, reasoning enabled setting);
a \textit{concise} instruction that asks the model to keep reasoning brief and ensure a final answer;
a \textit{token-budget} instruction, inspired by TALE~\cite{hanTokenBudgetAwareLLMReasoning2025a}, that asks the model to use no more than $8k$ reasoning tokens before answering;
an \textit{aspects} instruction that discourages step-by-step reasoning and instead asks the model to briefly consider key implementation concerns such as inputs, outputs, edge cases, and design choices;
and \textit{Plan-and-Solve}~\cite{wangPlanandSolvePromptingImproving2023}, which provides a contrasting structured-reasoning baseline without explicitly limiting reasoning length.
These conditions compare qualitative brevity, an explicit token limit, alternative reasoning structure, and structured planning against the model's default behaviour.
Each steering instruction is appended unchanged to every task prompt.
\textit{Results are shown in Fig.~\ref{fig:steering}.}

\paragraph{\textbf{Results}}

Prompt steering can reduce reasoning-trace overflow, but its effect is highly model-dependent.
The largest reduction is seen for \qwlil, where the \textit{aspects} prompt reduces average overflow from 21.4\% to 3.6\% while retaining similar accuracy.
\olmo\ also benefits from brevity-oriented prompts.
However, the models with the highest baseline overflow remain difficult to control: \llama\ and \opencr\ still overflow on more than 34\% and 45\% of tasks, respectively, even under their best prompt.
\openr\ behaves differently again, with every steering prompt increasing average overflow.

Accuracy is similarly mixed.
Most changes are modest, but \qwbig\ loses 17--19 percentage points under the qualitative \textit{concise} and \textit{aspects} prompts despite having little baseline overflow.
In contrast, the explicit \textit{token-budget} prompt largely preserves its accuracy.
More broadly, however, simply stating an $8k$ budget provides weak control: it substantially reduces the overall rate of reasoning beyond $8k$ only for \olmo, from 36\% to 22\%, while producing much smaller changes for most models and worsening \openr.
\textit{Plan-and-Solve} similarly provides no consistent reduction in overflow or improvement in accuracy.

The benchmark-level results show further variation.
On harder generation tasks, shorter or more structured reasoning can reduce overflow for \qwlil, \olmo, and \openr, and can sometimes improve accuracy at the same time.
However, the same interventions are much less reliable elsewhere.
Prompt steering therefore provides useful \emph{soft control}, but no single prompt consistently improves both accuracy and overflow across models and tasks.

\subsubsection{\textbf{Budgeted Transitions}}

\paragraph{\textbf{Set-up}}

Prompt steering can influence reasoning length, but even an explicit token-budget instruction cannot reliably enforce the requested limit.
We next test budgeted transitions, which directly control when reasoning ends.
Following prior work on budget forcing~\cite{muennighoffS1SimpleTesttime2025a}, we impose a transition budget \(B_t\) on the reasoning trace.
The model generates normally until either reasoning completes or \(B_t\) is reached.
If the model is still reasoning at \(B_t\), we append the end-of-reasoning token, \texttt{</think>}, and allow generation to continue in the answer segment.
We sweep \(B_t\) in $4k$-token steps from $0-28k$, and compare against the standard reasoning-enabled baseline.
Given the prevalence of loops among overflowed traces in RQ1, this also prevents repetitive reasoning from consuming the remainder of the output budget.
The overall $32k$-token output budget remains unchanged: tokens used for reasoning count toward this budget, and only the remaining budget is available for the final answer after transition.
\textit{Results are shown in Fig.~\ref{fig:transition}.}

\paragraph{\textbf{Results}}

Budgeted transitions provide more direct control, and the macro-average results show that many models benefit from being forced to answer after reasoning for a limited budget.
For most models, accuracy improves sharply within the first few thousand reasoning tokens and then begins to plateau.
 The strongest average effect is seen for \opencr, where a moderate transition budget improves \textit{pass@1} by almost 20 percentage points over standard reasoning.
\qwlil, \olmo, and \llama\ also exceed their reasoning-enabled baselines under suitable budgets, while the larger \texttt{Qwen} models recover or slightly exceed their baseline performance.
\openr\ is the clearest exception: immediate transition (reasoning disabled) performs best on average, suggesting that additional reasoning may be actively unhelpful for this model.
This reinforces that there is no single useful reasoning budget across models.

For budgeted transitions, the benchmark-level results show a clearer relationship with task difficulty.
On the more difficult generation benchmarks, \qwmid, \qwbig, \openr, and \opencr\ generally benefit from being allowed more reasoning before transition, with performance often continuing to improve at larger budgets.
In contrast, on the shorter \crux\ tasks, the \texttt{Qwen} models recover most of the benefit of reasoning within the first few thousand tokens, with little gained from substantially longer traces.
Other models remain more mixed, and some model--task pairs are harmed by additional reasoning altogether. This suggests that budgeted transitions provide stronger control than prompting, but the appropriate reasoning budget remains highly task- and model-dependent.

\begin{figure*}[t]
    \centering
    \includegraphics[
        width=\textwidth,
        alt={
Two grouped bar charts compare macro-average generation cost across
reasoning-disabled, best prompt-steering, best transition-budget, and
standard reasoning settings. Standard reasoning uses the most output
tokens for every model. Transition budgets substantially reduce both
output tokens and tokens per correct answer for several overflow-prone
models, including OLMo3-7B, Llama-R1-8B, OpenReason-7B, and
CodeReason-7B. Reasoning-disabled inference is the most token-efficient
setting for the Qwen models.
}]{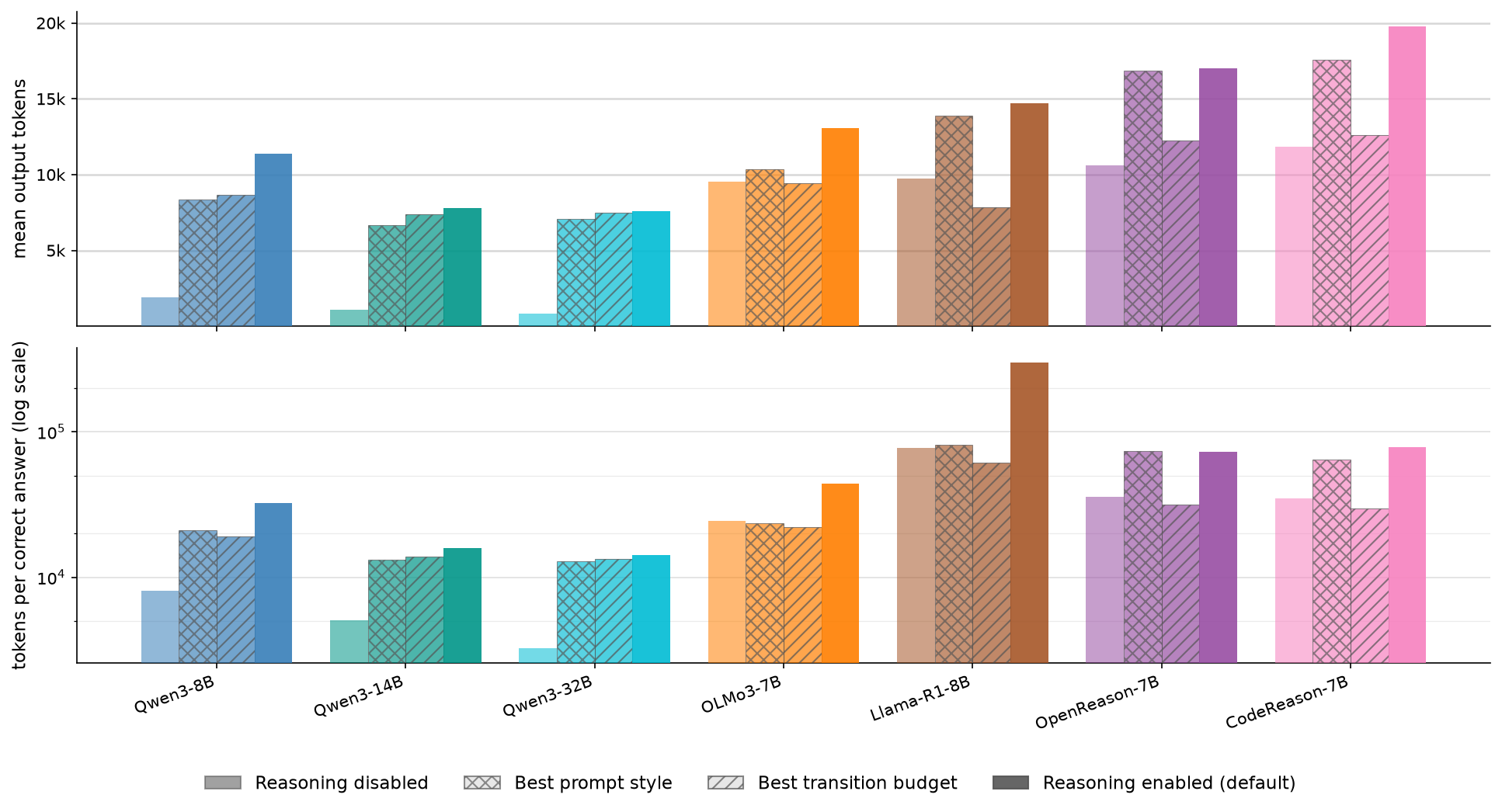}
    \caption{
        \textbf{\textit{Token Efficiency Analysis.}}
        Token efficiency across inference strategies, macro-averaged over all six benchmark settings.
        The top panel shows \textit{mean output tokens}, while the bottom panel shows \textit{tokens per correct answer} on a log scale.
        For each model, the prompt style and transition budget with the highest macro-average \textit{pass@1} are shown.
    }
    \label{fig:efficiency}

\end{figure*}

\smallskip
\begin{summarybox}{RQ3 Summary.}
Prompt steering and budgeted transitions provide two complementary forms of reasoning control.
Prompt steering offers \emph{soft control}: it is cheap and easy to deploy, and can substantially reduce overflow for some models, but its effects are inconsistent and can harm accuracy.
Budgeted transitions offer \emph{hard control}: they directly enforce when reasoning ends and often improve performance for overflow-prone models, but require access to the decoding process and an appropriate transition budget.
Neither approach is universally best, reinforcing that effective reasoning control remains strongly dependent on the model and task.
\end{summarybox}


\section{Token Efficiency}

Explicit reasoning introduces a direct generation cost, but token count alone does not capture whether this cost is worthwhile: a more expensive response may still be preferable if reasoning substantially improves accuracy.
We therefore compare four settings: reasoning-disabled inference, standard reasoning-enabled inference, the best prompt-steering style from RQ3, and the best transition budget from RQ3.
For the latter two, ``best'' is selected using the highest macro-average \textit{pass@1}; these retrospectively selected settings provide an upper-bound comparison rather than a deployable budget-selection strategy.
We report both \textit{mean output tokens} and \textit{tokens per correct answer}, calculated as the total number of generated tokens divided by the number of correct responses.
Values are macro-averaged across the six benchmark settings.

Standard reasoning-enabled inference is the most expensive setting for every model, with the largest gaps appearing for models that were most prone to reasoning-trace overflow.
For example, \llama\ averages roughly 15k output tokens under default reasoning, compared with 7.9k under its best transition budget, while \opencr\ falls from roughly 20k to 13k.
Prompt steering sometimes reduces this cost, but usually only modestly.
For the \texttt{Qwen} models, reasoning-disabled inference is much cheaper: \qwlil, for example, falls from roughly 11k tokens under default reasoning to 1.9k when reasoning is disabled.
Overall, transition budgets provide the most consistent reduction in output length while retaining strong accuracy.

The same pattern becomes more informative when we normalise by correctness.
For \olmo, \llama, \openr, and \opencr, the best transition budget requires substantially fewer tokens per correct answer than unconstrained reasoning, showing that longer reasoning does not provide proportional gains in accuracy.
For \llama, this falls from roughly 300k to 61k tokens per correct answer.
Its unusually high default cost is driven by the combination of long generations and relatively low baseline accuracy ($\approx$29\% \textit{pass@1}), rather than token length alone.
Prompt steering can also improve efficiency, but less reliably and generally by a smaller margin.
For the \texttt{Qwen} models, reasoning-disabled inference remains the most token-efficient setting overall; \qwlil\ falls from roughly 32k tokens per correct answer under default reasoning to 8.1k, although RQ2 showed that explicit reasoning can still provide substantially higher absolute accuracy on harder tasks.

Token efficiency therefore reinforces the same trade-off seen throughout the paper.
More reasoning can improve correctness, but its value depends strongly on the model.
For overflow-prone models in particular, controlling when reasoning ends can substantially increase the number of useful solutions obtained for a fixed token budget.


\section{Discussion}

\paragraph{\textbf{Reasoning is not free}}
Our results suggest that explicit reasoning should not be treated as an automatic improvement for code generation.
For some models and benchmarks, reasoning clearly improves \textit{pass@1}, especially on harder algorithmic tasks.
However, on simpler generation tasks, and for several model families, reasoning provides little benefit while increasing the risk of slow failure through reasoning-trace overflow.
This makes the usefulness of reasoning highly conditional: it depends on the model, the task, and whether the model can stop reasoning in time to produce code.

\paragraph{\textbf{Overflow is a practical failure mode}}
Reasoning-trace overflow is more than a token-count artifact.
Most complete overflow in our study is associated with detected reasoning loops, meaning models can spend substantial inference budget repeating reasoning without ever producing code.
However, long non-looping reasoning also remains common, showing that the broader problem is not reducible to looping alone.
From a user's perspective, both behaviours can waste generation budget and inference compute without producing a proportionate benefit.

\paragraph{\textbf{Code generation needs reasoning control}}
Our results point toward a middle ground between always reasoning and never reasoning.
When reasoning rarely overflows and provides clear accuracy gains, leaving it unconstrained can be worthwhile.
For overflow-prone models, prompt steering offers cheap \emph{soft control}, but remains inconsistent, while budgeted transitions provide more reliable \emph{hard control} when the decoding process is accessible.
The appropriate choice still depends on the model and task.



\section{Threats to Validity}

\paragraph{\textbf{Construct validity}}
We define reasoning-trace overflow using each model's explicit reasoning delimiters, and therefore rely on models following their expected output protocol.
This matches the practical failure we study---the model never transitions to a usable answer---but may not capture every possible formatting failure.
We measure generation cost using output tokens rather than wall-clock latency or energy, and token counts are most directly comparable within the same model because tokenizers differ across families.
Our loop detector targets sustained periodic repetition and may miss shorter or semantic loops, so the reported loop rates should be interpreted conservatively.

\paragraph{\textbf{Internal validity}}
We use greedy decoding and a fixed $32k$ output budget to provide controlled comparisons across models and tasks.
Different decoding strategies may change reasoning length, looping behaviour, and accuracy, while smaller operational budgets would increase the absolute rate of overflow.
Our prompt interventions are deliberately simple, and the ``best'' prompt styles and transition budgets are selected retrospectively using macro-average \textit{pass@1}, rather than by an adaptive deployment policy.

\paragraph{\textbf{External validity}}
We focus on relatively small open-weight models with accessible reasoning traces and on single-shot code-generation and code-reasoning benchmarks.
Larger or closed models, hidden-reasoning systems, and agentic software-engineering settings may behave differently.
In particular, multi-turn systems introduce additional factors such as tool use, retrieval, and repository context that are outside the scope of this study.
Data contamination is a further threat.
Older benchmarks may overlap with model pre-training data, but this is unlikely to inflate overflow rates: familiar or memorised solutions may require less search and planning, potentially making our estimates conservative.

\paragraph{\textbf{Ethics}} Finally, our study uses public benchmarks and open-weight models and does not involve human participants or personal data; we report accuracy alongside efficiency to avoid treating shorter reasoning as beneficial when it harms correctness.


\section{Conclusion}

Explicit reasoning can improve code generation, but our results show that it is not a universal benefit.
Across seven relatively small open-weight reasoning models and six code benchmark settings, we find that models often spend substantial portions of a $32k$ output budget on reasoning, sometimes failing to produce an answer at all.
Most complete overflow is associated with detected reasoning loops, although substantial long reasoning also occurs without looping.
This reasoning-trace overflow is most common on harder code-generation tasks, but remains highly model-dependent.
We further show that disabling explicit reasoning can be competitive when overflow is likely, while prompt steering and budgeted transitions offer complementary soft and hard forms of control, with transition budgets also improving token efficiency for several overflow-prone models.
Overall, our findings suggest that code-generation systems should treat reasoning as a controllable resource, rather than an always-on capability.

\newpage






\balance
\bibliographystyle{IEEEtran}
\bibliography{main}

\end{document}